\documentclass[letterpaper, 10 pt, conference]{icra/ieeeconf}  %

\IEEEoverridecommandlockouts                              %

\usepackage{cuted}
\usepackage{caption}

\usepackage{cite}

\usepackage{times}
\usepackage[utf8]{inputenc} %
\usepackage[T1]{fontenc}    %
\usepackage{url}            %
\usepackage{booktabs}       %
\usepackage{nicefrac}       %
\usepackage{microtype}      %
\usepackage{adjustbox}      %
\usepackage{graphics,color}
\usepackage{siunitx}        %
\usepackage{wrapfig}        %

\usepackage{algorithm}
\usepackage[noend]{algpseudocode}
\algrenewcommand\algorithmicindent{1em}
\algrenewcommand{\algorithmiccomment}[1]{%
\bgroup\hskip2em\textcolor{ourdarkgreen}{//~\textsl{#1}}\egroup}

\usepackage{listings}  %

\usepackage{amsfonts}       %
\usepackage{amsmath}       %
\usepackage{amssymb}
\usepackage{xspace}

\usepackage{xr-hyper}
\usepackage[hidelinks]{hyperref}%
\usepackage[capitalise, noabbrev]{cleveref} %

\makeatletter
\newcommand*{\addFileDependency}[1]{%
  \typeout{(#1)}
  \@addtofilelist{#1}
  \IfFileExists{#1}{}{\typeout{No file #1.}}
}
\makeatother

\usepackage[table]{xcolor}
\definecolor{ourblue}{rgb}{0.368,0.507,0.71}    %
\definecolor{ourdarkblue}{HTML}{354d72}         
\definecolor{ourlightblue}{HTML}{b4c5dc}    

\definecolor{ourorange}{rgb}{0.881,0.611,0.142} %
\definecolor{ourdarkorange}{HTML}{8f6213}         
\definecolor{ourlightorange}{HTML}{efc785}    

\definecolor{ourgreen}{rgb}{0.56,0.692,0.195}   %
\definecolor{ourdarkgreen}{HTML}{677f24}         
\definecolor{ourlightgreen}{HTML}{cee298}  

\definecolor{ourred}{rgb}{0.923,0.386,0.209}    %
\definecolor{ourdarkred}{HTML}{94300f}         
\definecolor{ourlightred}{HTML}{f7c5b2}  

\definecolor{ourviolet}{rgb}{0.528,0.471,0.701} %
\definecolor{ourdarkviolet}{HTML}{463b68}         
\definecolor{ourlightviolet}{HTML}{bcb3d4}  

\definecolor{ourbrown}{rgb}{0.772,0.432,0.102}  %
\definecolor{ourdarkbrown}{HTML}{905113}         
\definecolor{ourlightbrown}{HTML}{f1c093}  

\definecolor{ourazure}{rgb}{0.364,0.619,0.782}  %
\definecolor{ourdarkazure}{HTML}{2a5b79}         
\definecolor{ourlightazure}{HTML}{a6cae0}  

\definecolor{ourolive}{rgb}{0.572,0.586,0.}     %
\definecolor{ourdarkolive}{HTML}{5b5c00}         
\definecolor{ourlightolive}{HTML}{e3e395}  

\definecolor{ourgray}{RGB}{102,88,84}           %
\definecolor{ourdarkgray}{HTML}{362e2c}         
\definecolor{ourlightgray}{HTML}{bcb1b0}

\definecolor{ourblue2}{RGB}{9,134,223} %
\definecolor{ourdarkblue2}{RGB}{5,97,164} %
\definecolor{ourlightblue2}{RGB}{132,201,250} %

\definecolor{ourorange2}{RGB}{224,90,18} %
\definecolor{ourdarkorange2}{RGB}{160,63,9} %
\definecolor{ourlightorange2}{RGB}{246,175,137} %

\definecolor{ouryellow2}{RGB}{227,213,25} %
\definecolor{ourdarkyellow2}{RGB}{177,166,17} %
\definecolor{ourlightyellow2}{RGB}{242,235,140} %

\definecolor{ourpink2}{RGB}{247,24,139} %
\definecolor{ourdarkpink2}{RGB}{164,4,86} %
\definecolor{ourlightpink2}{RGB}{250,163,207} %

\definecolor{ourgreen2}{RGB}{159,198,52} %
\definecolor{ourdarkgreen2}{RGB}{109,138,30} %
\definecolor{ourlightgreen2}{RGB}{209,228,154} %

\definecolor{ourgray2}{RGB}{124,124,115} %
\definecolor{ourdarkgray2}{RGB}{87,87,81} %
\definecolor{ourlightgray2}{RGB}{194,194,189} %

\definecolor{ourhighlight}{HTML}{d4e1f5}
\definecolor{ourhighlight2}{HTML}{7ea6e0}

\definecolor{gray}{rgb}{0.5,0.5,0.5}

\lstdefinelanguage{prompt}{
    basicstyle=\ttfamily\small\color{black},
    keywordstyle=\color{black},
    commentstyle=\color{black},
    identifierstyle=\color{black},
    stringstyle=\color{black},
    keywords={}, 
    morekeywords={},
    literate={},
    aboveskip=\bigskipamount,
}

\newcommand{\method}{NaviTrace\xspace}

\makeatletter
\DeclareRobustCommand\onedot{\futurelet\@let@token\@onedot}
\def\@onedot{\ifx\@let@token.\else.\null\fi\xspace}
\makeatother

\newcommand{\eg}{e.g\onedot}

\newcommand{\greencheck}{{\large\color{ourgreen}{$\checkmark$}}}
\newcommand{\redcross}{{\Large\color{ourred}{$\pmb\times$}}}
\newcommand{\human}{\includegraphics[height=0.9em]{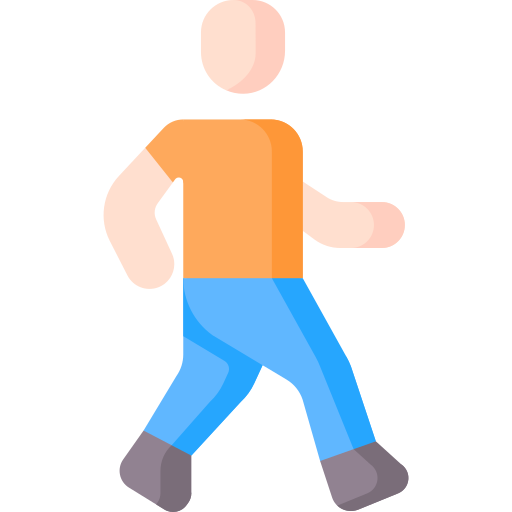}}
\newcommand{\legged}{\includegraphics[height=0.9em]{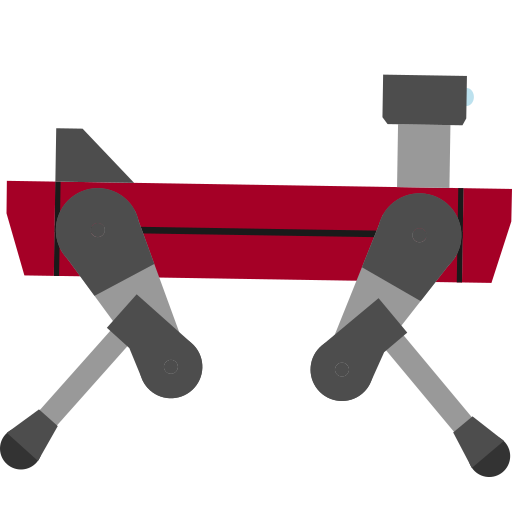}}
\newcommand{\wheeled}{\includegraphics[height=0.9em]{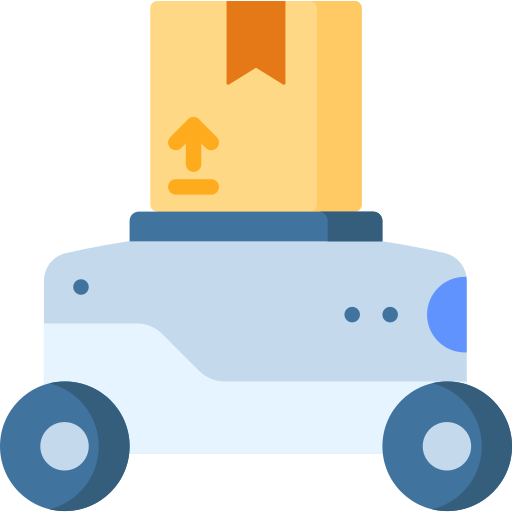}}
\newcommand{\bicycle}{\includegraphics[height=0.9em]{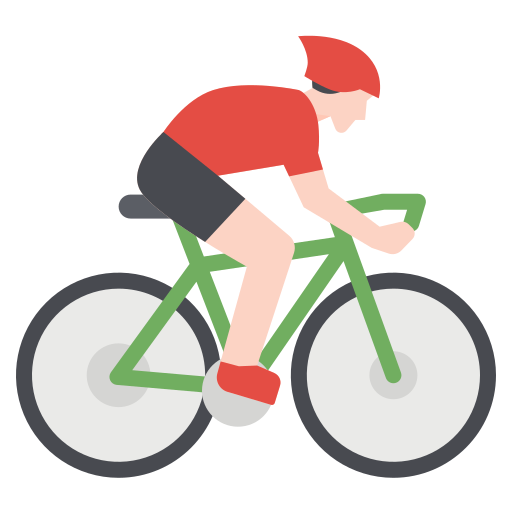}}
\newcommand{\real}{\includegraphics[height=0.9em]{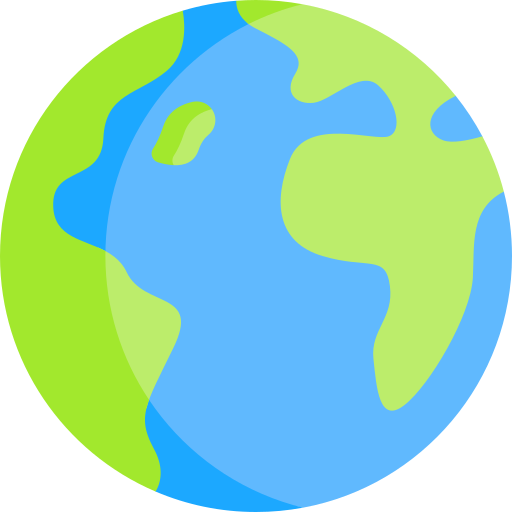}}
\newcommand{\simulation}{\includegraphics[height=0.9em]{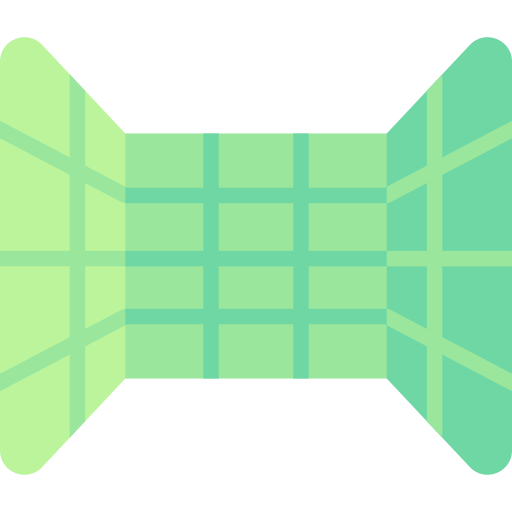}}
\newcommand{\manual}{\includegraphics[height=0.9em]{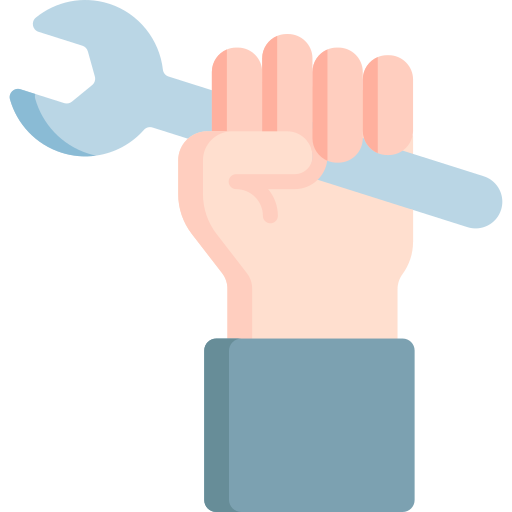}}
\newcommand{\automatic}{\includegraphics[height=0.9em]{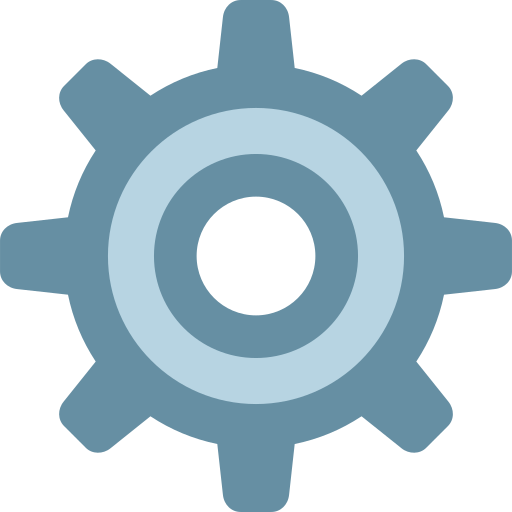}}

\newif\ifanonymous

\graphicspath{{graphics/}}

\usepackage{lineno}

\usepackage{svg} %
\svgpath{{graphics/}}

\title{\LARGE \bf
\method: Evaluating Embodied Navigation of Vision-Language Models
}

\author{
Tim Windecker$^{1,2}$, Manthan Patel$^{1}$, Moritz Reuss$^{2}$, Richard Schwarzkopf$^{3}$, Cesar Cadena$^{1}$,\\ Rudolf Lioutikov$^{2,4}$, Marco Hutter$^{1}$ and Jonas Frey$^{1}$%
\thanks{$^{1}$Robotic Systems Lab, ETH Zurich, Zurich, Switzerland}
\thanks{$^{2}$Intuitive Robots Lab, KIT, Karlsruhe, Germany}
\thanks{$^{3}$FZI Research Center for Information Technology, Karlsruhe, Germany}
\thanks{$^{4}$Robotics Institute Germany}
}

\begin{document}

\maketitle

\begin{strip}
    \vspace{-1.5cm}
    \centering
    \includegraphics[width=\textwidth]{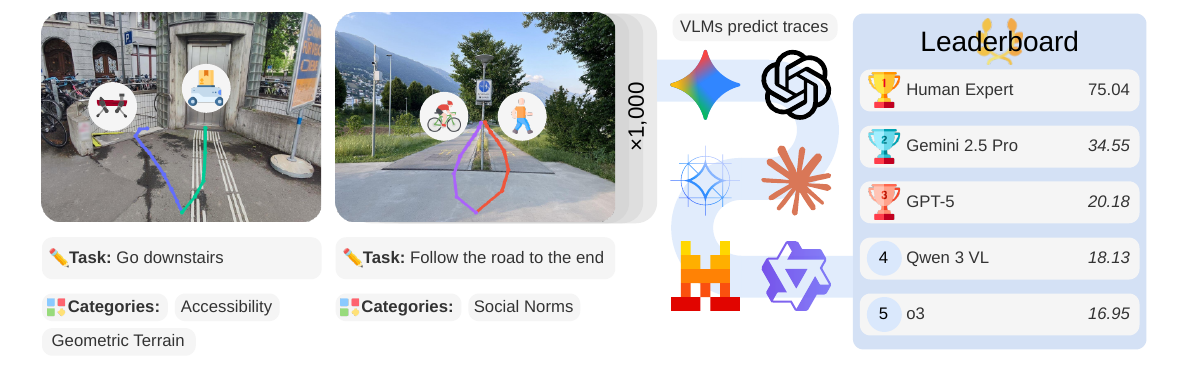}
    \captionof{figure}{We introduce \textbf{\method}, a novel VQA benchmark for VLMs that evaluates models on their embodiment-specific understanding of navigation across challenging real-world scenarios.}
    \label{fig:intro}
\end{strip}

\thispagestyle{empty}
\pagestyle{empty}

\begin{abstract}
Vision–language models demonstrate unprecedented performance and generalization across a wide range of tasks and scenarios.
Integrating these foundation models into robotic navigation systems opens pathways toward building general-purpose robots.
Yet, evaluating these models’ navigation capabilities remains constrained by costly real-world trials, overly simplified simulations, and limited benchmarks.
We introduce \method, a high-quality Visual Question Answering benchmark where a model receives an instruction and embodiment type (human, legged robot, wheeled robot, bicycle) and must output a 2D navigation trace in image space.
Across 1000 scenarios and more than 3000 expert traces, we systematically evaluate eight state-of-the-art VLMs using a newly introduced semantic-aware trace score. 
This metric combines Dynamic Time Warping distance, goal endpoint error, and embodiment-conditioned penalties derived from per-pixel semantics and correlates with human preferences. 
Our evaluation reveals consistent gap to human performance caused by poor spatial grounding and goal localization. 
\method establishes a scalable and reproducible benchmark for real-world robotic navigation.
The benchmark and leaderboard can be found at \url{https://leggedrobotics.github.io/navitrace_webpage/}.
\end{abstract}

\section{INTRODUCTION}\label{sec:intro}

The rise of foundation models with general-purpose capabilities has sparked a push to develop robots that are equally general-purpose---capable of flexible, wide-ranging behavior in real-world environments. 
Given their significant potential, it is crucial to rigorously assess how these models perform in real-world robotic applications.
Such evaluations inform model development, reveal limitations, and establish benchmarks for comparing approaches.
However, assessing the navigation capabilities of these models remains challenging.
Navigation is critical for numerous robotic applications, including last-mile delivery, industrial inspection, search and rescue, and assistive tools for visually impaired people.
Yet only limited evaluation methods currently exist for these capabilities.

\begin{table*}[tbp]
\centering
\caption{An overview of vision language navigation and VLM datasets.} \label{tab:bench}
\adjustbox{max width=\linewidth}{
\begin{tabular}{l p{5em} m{1.5em} m{1.5em} p{10em} m{3em} p{20em} m{5.5em} p{16em} m{5.5em}}
\toprule
& \multicolumn{3}{c}{Task} & \multicolumn{2}{c}{Data Source} & \multicolumn{2}{c}{Annotations} & & \\
    \cmidrule(lr){2-4}
    \cmidrule(lr){5-6}
    \cmidrule(lr){7-8}
Name & Type & VLM & Nav. & Description & Sim \simulation \, Real \real & Description & Automatic \automatic \, Manual \manual & Scoring & Embodiments \\
\midrule
R2R \cite{r2r} & VLN & \redcross & \greencheck & MP3D \cite{matterport3d} & \simulation & 21,567 language navigation instructions & \automatic\,/\manual & Success Rate, Navigation Error & \human \\
REVERIE \cite{reverie} & VLN & \redcross & \greencheck & MP3D \cite{matterport3d} & \simulation & 21,702 language instructions that require navigating and identifying an object & \automatic\,/\manual & Success Rate, SPL & \human \\
RxR \cite{rxr} & VLN & \redcross & \greencheck & MP3D \cite{matterport3d} & \simulation & 126k multilingual time-aligned language instructions, 126k demonstration paths & \automatic\,/\manual & Success Rate, SPL, Navigation Error, Normalized Dynamic Time Warping & \human \\
OctoNav-Bench \cite{octonav} & Embodied Navigation & \redcross & \greencheck & Habitat (MP3D, HM3D, Gibson, ProcTHOR) & \simulation & 45k+ annotated instructions with trajectories that combine the task types: object goal, point goal, image goal, instance-image goal, and VLN, 10k+ instruction-think-action pairs & \automatic & Success Rate, SPL & \human \\
EgoWalk \cite{egowalk} & VLN & \redcross & \greencheck & \SI{50}{h} of egocentric navigation recordings & \real & Automatic traversability region and language goals with extracted odometry trajectories & \automatic & MSE, Absolute Displacement Error, and Final Displacement Error & \human \\
CityWalker \cite{citywalker} & Point Goal Nav. & \redcross & \greencheck & \SI{2000}{h} of city walking videos from the internet & \real & With visual odometry extracted trajectory poses & \automatic & Average Orientation Error & \human \\
SocialNav-SUB \cite{socialnavsub} & Social Navigation VQA & \greencheck & \redcross & SCAND \cite{scand} & \real & 4968 unique questions, 24840 human responses & \automatic\,/\manual & Probability of Agreement, Consensus-Weighted Probability of Agreement & \legged, \wheeled \\
Social-LLaVA \cite{socialllava} & Social Navigation VQA & \greencheck & \redcross & SCAND \cite{scand} & \real & 40k questions fully annotated by humans & \manual & Human judges & \legged, \wheeled \\
\rowcolor{ourhighlight}\textbf{\method (ours)} & Nav. Traces for VLMs & \greencheck & \greencheck & 1000 diverse real images & \real & 1k language instructions, 3k+ traces that describe 2D paths & \manual & Human-preference aligned nav. metric & \human, \legged, \wheeled, \bicycle \\
\bottomrule
\end{tabular}
}
\end{table*}

Applying existing Vision Language Models (VLMs) to the navigation task itself can be straightforward: A text instruction and an image observation can be provided as input, prompting the language model to generate a text description that the robot’s control system can translate into motor commands~\cite{navila, uninavid, rynnbrain, abot-n0, socialnav}.
While various flavors of such systems exist, it remains unclear which VLM system performs best.

To evaluate this, there exist three main directions in the literature:
The first is to perform real-world closed-loop rollouts on a set of navigation tasks and measure the success rate. 
However, such experiments are expensive, time-consuming, and inherently do not scale well for evaluating performance across diverse operating environments, while also lacking reproducibility.
The second is to run the same closed-loop experiments in simulation. This approach improves reproducibility but still faces significant limitations. The diversity of tasks is constrained by the scenarios created in simulation, which are inherently simplified in terms of dynamics, mostly feature static scenes, and have limited semantics.
Important factors, such as varying terrain properties or social norms, that should influence an agent’s navigation behavior, are difficult to encode.
Lastly, Visual Question Answering (VQA) benchmarks can overcome some of these limitations by making use of high-quality human annotations, which can incorporate semantics, social preferences, and geometric cues.
While several navigation-focused VQA datasets exist, they typically (i) constrain outputs to text answers rather than trace-level plans, and (ii) evaluate only legged or wheeled robot embodiments~\cite{socialnavsub, socialllava}.

To fill this gap, we introduce \method, a VQA benchmark specially designed to evaluate embodiment-specific navigation performance across 1,000 diverse scenarios and four different embodiments.
Each task within \method consists of a single real-world image paired with a high-quality language instruction, enabling efficient data collection while capturing challenging navigation tasks.
Following the most intuitive approach to answering navigation questions, we provide solutions per embodiment as 2D paths in image space, which we refer to as traces.
This carefully chosen formulation is more expressive than low-level commands such as “Forward”~\cite{navila} and can also support longer-horizon planning.
It can be seen as an extension of pointing---a common task that is evaluated and optimized in current foundation models~\cite{geminirobotics, molmo} and widely used to assess the visual grounding of VLMs~\cite{pointarena}.
Furthermore, traces have proven beneficial for addressing manipulation tasks~\cite{magma, llarva, scaffolding, tracevla}.
\method tests VLMs for instruction following, spatial understanding, and physical understanding of varying embodiments (human, legged robot, wheeled robot, and bicycle), and categorizes scenarios based on the type of navigation challenges.

We develop a semantic-aware score to measure how well the predicted navigation trace aligns with human preferences.
To achieve this, we combine the Dynamic Time Warping distance to a ground-truth trace, goal endpoint error, and pixelwise embodiment-conditioned penalties derived from a semantic segmentation model.
We show that our metric, while inexpensive to compute and annotate, is competitive with more expensive human-derived metrics in aligning with human preferences.

Specifically, our main contributions are:

\begin{enumerate}
    \item \textbf{\method:} A novel high-quality benchmark for evaluating the ability of VLMs to predict how different embodiments navigate in 1000 diverse and challenging real-world scenarios.
    \item \textbf{Semantic-aware Score}: A new metric to measure the accuracy of 2D traces for real-world images. We test the score for alignment with human preferences by showcasing its correlation to human expert judgments.
    \item \textbf{Evaluation of VLMs:} Comprehensive assessment of current state-of-the-art VLMs on our benchmark.
\end{enumerate}

\section{RELATED WORK}\label{sec:rel_work}

\cref{tab:bench} provides an overview of benchmarks evaluating the navigation performance of vision-based agents and VLMs.

\textbf{Vision Language Navigation Benchmarks.}
Several relevant works focus on the evaluation of vision-language navigation (VLN) tasks, where agents follow natural language instructions using visual input.
Well-established benchmarks include Room-to-Room (R2R)~\cite{r2r}, REVERIE~\cite{reverie}, and Room-Across-Room (RxR)~\cite{rxr}.
R2R and RxR feature fine-grained instructions, while REVERIE uses coarser descriptions that also require object identification (\eg, “Bring me the bottom picture next to the top of the stairs on level one”~\cite{reverie}).
All three benchmarks rely on Matterport3D (MP3D) scenes~\cite{matterport3d} in simulation, which provides realistic indoor environments but restricts navigation to discrete viewpoint transitions without realistic physics simulation.
OctoNav-Bench~\cite{octonav} extends VLN benchmarks by combining multiple task types into free-form instructions.
It leverages the Habitat simulator~\cite{habitat} that supports continuous action spaces.
While simulators enable the training of reinforcement learning policies, they remain constrained to the underlying indoor training environments and often fail to accurately model the physical interactions corresponding to visual observations, which is one of the main causes of the visual sim-to-real gap.
Other benchmarks address this limitation by directly collecting real-world data.
EgoWalk~\cite{egowalk} records egocentric navigation with annotated language goals and extracted trajectories.
CityWalker~\cite{citywalker} uses internet videos and visual odometry to extract trajectories, and therefore does not have language-conditioned tasks.

Existing VLN benchmarks present several limitations for VLM evaluation.
Most require trajectory predictions in specialized action spaces that VLMs cannot natively predict. 
They focus exclusively on human navigation, overlooking cross-embodiment challenges. 
To address these gaps, \method uses manually collected real-world images and VLM-accessible 2D trace prediction across multiple embodiment types.

\begin{figure*}[htbp]
  \centering
  \includegraphics[width=0.7\textwidth]{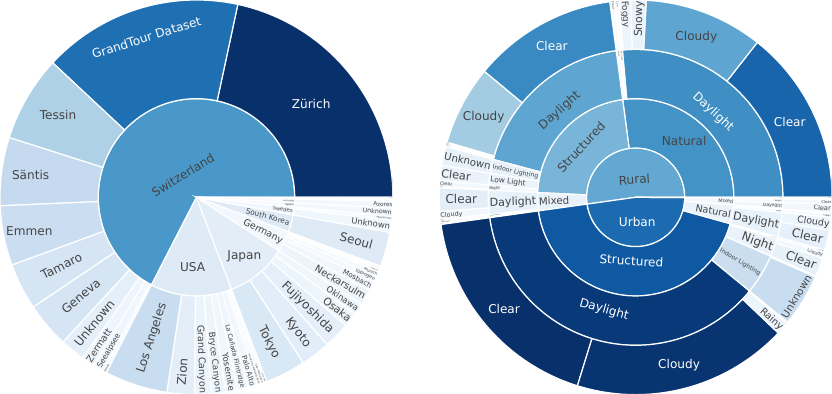}
  \caption{\textbf{Left:} Geographic distribution of image sources, with the inner circle denoting countries and the outer circle specifying cities or regions.
  Images originating from the GrandTour Dataset~\cite{grandtour} are explicitly marked in the outer circle.
  \textbf{Right:} Distribution of scenarios by setting (urban vs.~rural), environment type (natural vs.~structured), lighting, and weather.}
  \label{fig:data_div}
\end{figure*}

\textbf{Vision-Language Model Benchmarks.}
There exists a variety of VLM benchmarks, which adapt standard computer vision tasks into VLM-compatible formats~\cite{blink,cambrian,gpt4o_vision}, test embodied skills such as pointing~\cite{pointarena}, spatial understanding~\cite{embspatial}, or embodied reasoning~\cite{geminirobotics}.
Most relevant to \method are benchmarks in social navigation.
SocialNav-Sub~\cite{socialnavsub} evaluates models through VQA on videos of robot-human interactions, asking about spatial relations, robot and pedestrian motion, as well as interaction dynamics.
Similarly, the VQA dataset SNEI~\cite{socialllava} contains social scenarios in crowded spaces, asking models to describe perceptions, predict future movements, reason about robot actions, and give a general explanation of what is happening.
However, to the best of our knowledge, no existing benchmark directly evaluates VLMs on navigation tasks or their understanding of differences between embodiments when navigating. 

\section{NAVITRACE BENCHMARK}\label{sec:bench}

We introduce \method, a benchmark for evaluating the ability of VLMs to predict navigation strategies for different embodiments in real-world scenarios (see \cref{fig:intro}).
To ensure relevance, diversity, and high-quality annotation, we manually collect real-world images and perform all labeling by hand.
The dataset contains 1{,}000 scenarios with more than 3{,}000 traces, divided evenly into validation and test splits.
The test set annotations remain secret and are used to evaluate the public leaderboard.
During evaluation, a VLM receives a structured prompt with an image, a task description, and an embodiment type.
The model must predict a path that solves the task, and its output is measured using our novel task-specific score function.

\subsection{Data Collection} \label{sec:data_collection}

Each scenario in \method combines images, instructions, traces, and embodiment types to capture realistic navigation challenges (see \cref{fig:intro} for examples).

\textbf{Image.}
Each scenario includes a distinct first-person image of a real-world environment.
Most images are crowd-sourced and captured with consumer devices such as phones or GoPros, complemented by $164$ curated samples from the publicly available GrandTour dataset~\cite{grandtour}.
To preserve privacy, we anonymize all personal data using EgoBlur~\cite{egoblur} to blur faces and license plates.

\textbf{Task Instruction.}
Each image is paired with a manually written instruction solvable purely from the visual information.
These instructions emphasize cases where different embodiments behave differently, while still reflecting everyday scenarios.
They are formulated either as goals (\eg, "Go to the red car") or as directional instructions (\eg, "Go forward, then turn left at the traffic light.").

\textbf{Task Categories.}
To classify capabilities of models according to navigation-relevant attributes,  we tagged each scenario with one or more categories, describing the main challenges of the navigation task:
\begin{itemize}
    \item \textbf{Geometric Terrain Property Assessment:} Decisions based on the shape, structure, or 3D geometry of permanent terrain features (\eg, stairs, a cliff, or closed doors).
    \item \textbf{Semantic Terrain Property Assessment:} Decisions requiring semantic understanding of properties (\eg, sidewalk, or road), or physical qualities (\eg, terrain stiffness, or friction).
    \item \textbf{Accessibility:} Barrier-free access for embodiments such as wheelchairs or delivery robots (\eg, wheelchair ramps, or automatically opening doors).
    \item \textbf{Visibility:} Scenarios with occlusions, poor lighting, or ambiguous information (\eg, blocked lines-of-sight, or unclear signage).
    \item \textbf{Social Norms:} Normative constraints from rules or signage (\eg, crosswalks, walking on a pedestrian walkway, or following a sign to not step on grass).
    \item \textbf{Dynamic Obstacle Avoidance:} Reacting to and planning around moving obstacles (\eg, humans, or vehicles).
    \item \textbf{Stationary Obstacle Avoidance:} Navigation around fixed obstacles not part of the general terrain structure (\eg, debris, or road closures).
\end{itemize}

\textbf{Ground-Truth Trace.}
We define a trace as a sequence of 2D points given as image coordinates that describes a navigation path.
This representation is detached from robot-specific controls, ensuring compatibility with diverse model architectures.
We draw one trace per suitable embodiment and multiple traces if there are equally valid and fast alternatives (\eg, avoiding an obstacle from the left or right).

\textbf{Embodiments.}
We model four embodiment types to capture various real-world navigation behaviors:
\begin{itemize}
\item \textbf{Human:} A regular pedestrian unable to climb tall obstacles.
\item \textbf{Legged Robot:} A quadruped (\eg, ANYmal~\cite{anymal}) with behavior similar to humans but shorter in stature.
\item \textbf{Wheeled Robot:} A small, wheelchair-like delivery robot that favors walkways and ramps.
\item \textbf{Bicycle:} A cyclist following traffic rules, preferring bike lanes or streets, and avoiding stairs.
\end{itemize}
We deliberately exclude cars, since their viewpoint differs fundamentally from the embodiments above.

\subsection{Data Quality}

To ensure scenario diversity, we analyze the dataset along five factors: (i) geographical location, (ii) urban vs.\ rural setting, (iii) natural vs.\ structured environment, (iv) lighting conditions, and (v) weather.
The geographic distribution is shown on the left in \cref{fig:data_div}.
While the dataset is geographically concentrated in Switzerland, it also includes samples from several other countries to provide broader international representation.

The right side of \cref{fig:data_div} summarizes the distribution of scenarios across the remaining factors.
The scenarios are balanced between an urban and rural setting.
Structured environments appear more frequently than natural ones, because urban scenes rarely contain natural elements.
Most images were captured in daylight under clear or cloudy weather, resulting in high visual quality.
This shows a tendency toward favorable conditions for vision, however the benchmark primarily targets navigation challenges rather than visual perception under difficult conditions.

\subsection{Score} \label{sec:score}

To fairly evaluate VLM-generated navigation traces, we design a score function that balances three factors: (i) how closely the path follows the ground truth, (ii) whether it reaches the intended goal, and (iii) whether it avoids unsafe or irrelevant regions.
Later, we describe how we make the score range easier to interpret and show that our score formulation aligns well with human preferences.
Formally, a trace is a sequence of points $T=[(x_1, y_1), \dots, (x_n, y_n)]$ in image pixel space.
We compare it against ground-truth traces across modalities $T'=[(x'_1, y'_1), \dots, (x'_m, y'_m)] \in \mathcal{G}$ and select the trace with the lowest error:
\begin{align}
\textrm{Score}(T, \mathcal{G}) = \underset{T' \in \mathcal{G}}{\min}\,& \textrm{DTW}(T, T') + \textrm{FDE}(T, T')
\\\nonumber
&+ \textrm{Penalty}(T)
\end{align}

\textbf{Trace Similarity:}
We utilize Dynamic Time Warping (DTW)~\cite{dtw} with the Euclidean distance as the error metric, to measure trace similarity.
DTW aligns sequences by stretching or compressing the time axis and can be computed using dynamic programming:
\begin{align}
DTW(T,T') =&\ D(n,m)
\\
D(0,0) =&\ 0
\\
D(i,0) =&\ D(0,j) = \infty \;\; (i,j>0)
\\
D(i,j) =&\ d((x_i, y_i), (x'_j, y'_j))
\\\nonumber
&+ \min\{D(i-1,j),\; D(i,j-1), 
\\\nonumber
 &\ D(i-1,j-1)\}
\end{align}

\textbf{Goal Reaching:} To reward reaching the correct target, we add the Final Displacement Error (FDE), which measures the Euclidean endpoint distance:
\begin{align}
    \textrm{FDE}(T, T') =&\ \textrm{d}((x_n, y_n), (x'_m, y'_m))
\end{align}

\begin{figure*}
  \centering
  \includegraphics[width=0.88\textwidth]{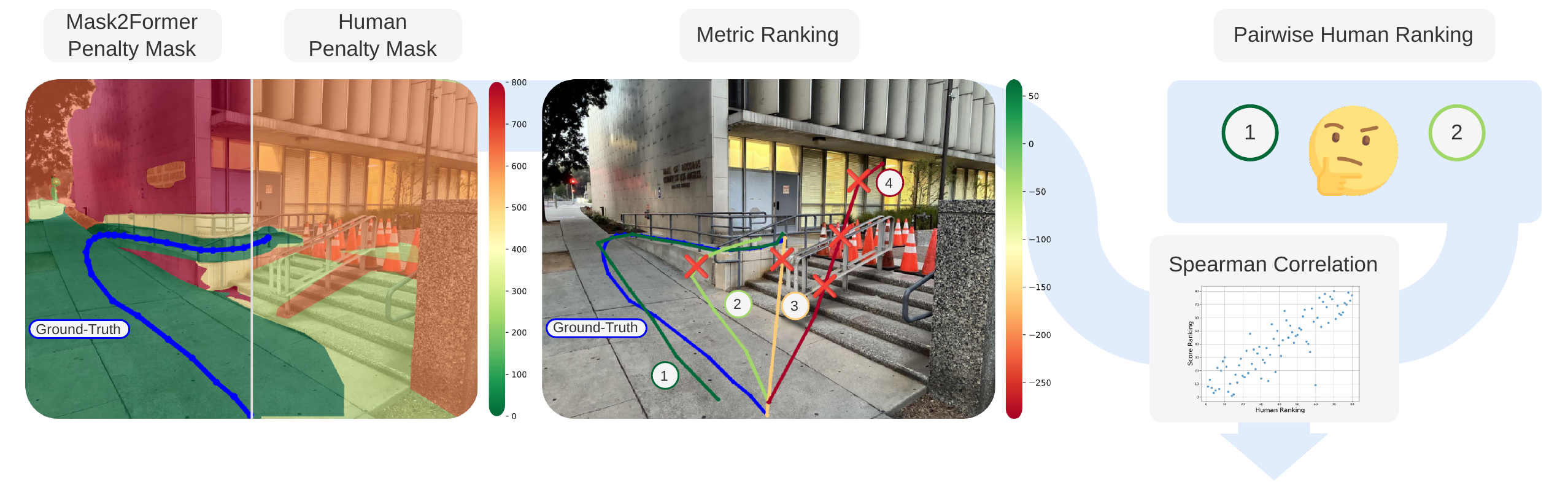}
  \caption{\textbf{Left:} Comparison between penalty cost masks based on Mask2Former and manual segmentation. These masks are used to punish traces crossing unsafe or irrelevant areas. \textbf{Right:} We show that the score function aligns with human preference by calculating the correlation between the score ranking and a pairwise ranking created by a human.}
  \label{fig:metric_trace_ranking}
\end{figure*}

\textbf{Semantic Penalty:}
Finally, we introduce embodiment-specific semantic costs that penalize traces crossing undesired regions.
Using a Mask2Former model~\cite{mask2former} trained on Mapillary Vistas~\cite{mapillary}, we infer semantic masks and map each class to manually tuned penalty values $\textrm{m}_{e}(S_i)$ depending on embodiment $e$.
Classes representing more dangerous areas or obstacles are assigned higher penalty values.
To allow for small deviations, we exclude a tolerance band around the ground-truth.
The penalty is averaged pixel-wise along the predicted trace:
\begin{align}
\textrm{Penalty}(T) =& \frac{1}{|\textrm{Pixels}(T)|} \sum_{i \in \textrm{Pixels}(T)} \textrm{m}_{e}(S_i)
\end{align}

\textbf{Scaling:}
In order to make the score values easier to interpret, we scale them to a range where the worst score is at $0$ and the best score at $100$.
We achieve this by setting the ground-truth performance $0$ as the lower bound.
For the upper bound, we select the performance of just drawing a vertical line through the image center, which corresponds to the Straight Forward baseline performance of $3234.75$.
This results in the scaled score function:
\begin{align}
\widehat{\textrm{Score}}(T, \mathcal{G}) = \frac{3234.75 - \textrm{Score}(T, \mathcal{G})}{3234.75} \cdot 100
\end{align}
Note that negative values are possible and do occur in our later experiments as some models perform worse than the Straight Forward baseline.

\textbf{Evaluation:} We acknowledge that, for each term, within our proposed score function, different choices may be made.
Complexities can arise from defining the exact start and end points, accounting for whether the path intersects hazardous terrain, and recognizing that distances measured in 2D image space do not directly translate to 3D navigation behavior.
For example, at greater distances, higher accuracy is required to follow a path, although in practice, a receding-horizon control approach may be applied.

Given these complexities and challenges, it is essential to evaluate whether the proposed score function aligns with human judgments.
To do so, we compute Spearman correlations~\cite{spearmancorr} between a human pairwise ranking of predictions and several score variants.
To cover the full quality range of predictions, we compile an equal mix of human, model, baseline, and intentionally flawed predictions.
Annotators perform pairwise comparisons to produce a human ranking, which we correlate with each score variant using Spearman’s rank correlation.
This correlation ranges from perfect agreement ($\pm1$) to no relation ($0$), ignoring linear relationships in favor of rank order.
The procedure is illustrated in \cref{fig:metric_trace_ranking}.

\begin{table}[htbp]
\centering
\caption{Spearman correlation between variants of the score function and human ranking.} \label{tab:metric}
\adjustbox{max width=\columnwidth}{
\begin{tabular}{p{15em} c}
\toprule
Score Variant & Spearman Correlation [$\uparrow$] \\
\midrule
RMSE & 0.8167 \\
Fréchet & 0.8310 \\
DTW & 0.8417 \\ 
DTW + FDE & 0.8656\\
DTW + FDE + Manual Penalty & \textbf{0.8723} \\
DTW + FDE + Mask2Former \textbf{(ours)} & 0.8707 \\
\bottomrule
\end{tabular}
}
\end{table}

We begin by comparing plain DTW similarity with alternative measures such as root mean square error (RMSE) and discrete Fréchet distance (see \cref{tab:metric}), and observe that DTW consistently achieves the highest performance across all three trace-similarity metrics.
We then evaluate the additional contribution of the FDE term for goal-reaching and find a further, consistent improvement in performance.
Extending the score with our Mask2Former-based semantic penalty leads to another clear performance gain.
To assess how well these automatically derived semantic cost terms align with expert annotations, we asked human annotators to semantically segment images into task-relevant, irrelevant (but safe), and hazardous regions (see \cref{fig:metric_trace_ranking}).
While dense manual semantic labeling is resource-intensive, it offers only limited performance gains over our Mask2Former-based strategy.
Taken together, these findings validate our decision to combine all three terms.

\section{EXPERIMENTS}\label{sec:exp}

Our experiments aim to address three key questions:
\begin{enumerate}
\item How well do current VLMs predict navigation traces?
\item Does performance vary with embodiment or task category?
\item Which aspects of the tasks pose the greatest challenges?
\end{enumerate}
To answer these questions, we first establish five baselines that give insight into the core difficulties of predicting navigation traces.
Next, we outline our deployment of state-of-the-art VLMs, before presenting and analyzing the benchmark results for the test split.

\subsection{Baselines} \label{sec:baselines}

\begin{figure*}[htbp]
  \centering
  \includegraphics[width=\textwidth]{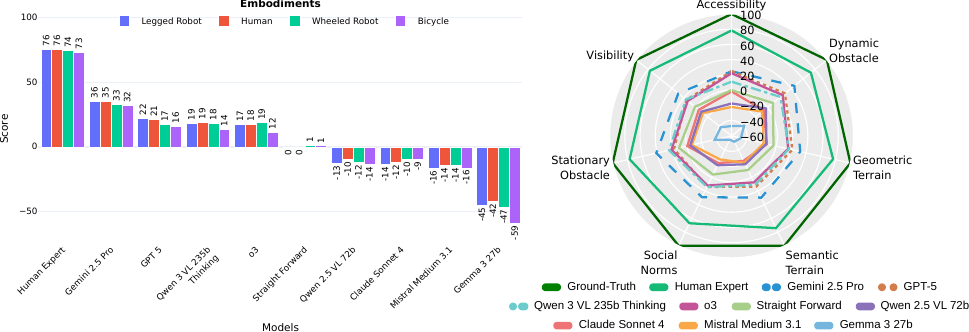}
  \caption{\textbf{Left:} Ranking of VLMs, the uninformed baseline Straight Forward, and human expert performance split into each embodiment. Note that a higher score is better. \textbf{Right:} Performance per task category for the same models.}
  \label{fig:bench_res}
\end{figure*}

We compare VLM performance against five baselines:
\begin{itemize}
    \item \textbf{Human:} Multiple participants collectively solve all test split scenarios, providing an upper bound for model performance.
    \item \textbf{Straight Forward:} Places a vertical line through the image center.
    \item \textbf{Oracle-Goal Straight Line:} Connects the given start and goal points with a direct line. In contrast to VLMs, this method has direct access to the goal and start point.
    \item \textbf{Only predict goal point:} To isolate the difficulty of identifying the goal, we use Gemini 2.5 Pro to predict only the goal location and connect it to the given start via a straight line.
    \item \textbf{Only predict path:} Conversely, given both start and goal, Gemini 2.5 Pro predicts only the navigation path.
\end{itemize}
Together, these baselines capture informed strategies, an upper bound with human performance, and provide context for assessing VLM performance.

\subsection{Models}

We evaluate all VLMs by querying each model through API calls.
After preliminary testing, we select five representative proprietary models: Gemini 2.5 Pro~\cite{gemini2.5}, GPT-5~\cite{gpt5}, o3~\cite{o3}, Claude Sonnet 4~\cite{claude4}, and Mistral Medium 3.1~\cite{mistral}.
We also include three open-weight models: Qwen 2.5 VL 72B~\cite{qwen2.5vl}, Qwen 3 VL 235B A22B Thinking ~\cite{qwen2.5vl}, and Gemma 3 27B~\cite{gemma}.
Among these models, Gemini 2.5 Pro, GPT-5, o3, and Qwen 3 VL automatically generate reasoning steps.
Each model receives a carefully crafted prompt specifying the task, output format, expected embodiment behavior, and embodiment type.
Models are instructed to return navigation traces as lists of normalized 2D points in JSON format, which we parse to compute performance scores.

\subsection{Performance}

\begin{figure*}[htbp]
  \centering
  \includegraphics[width=0.9\textwidth]{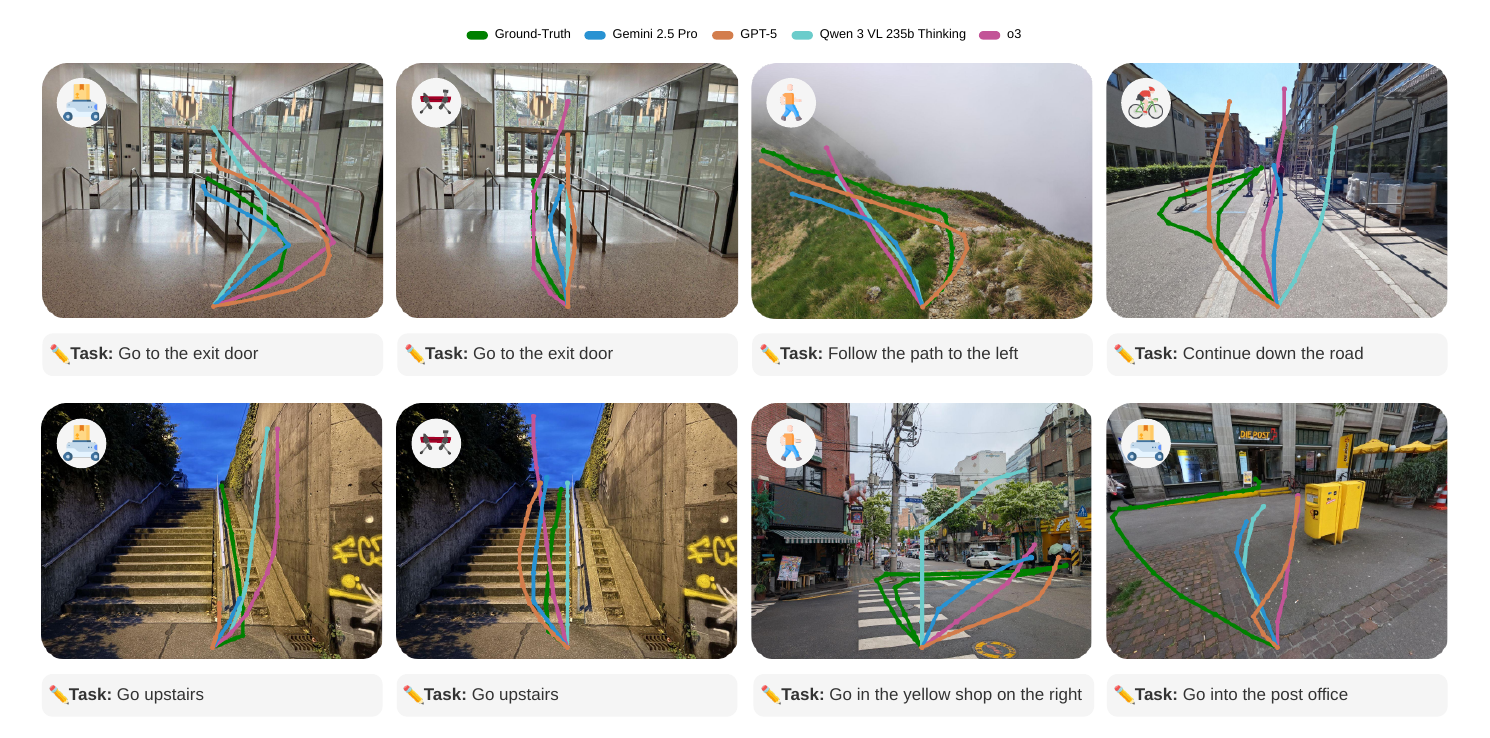}
  \caption{Example predictions by the models Gemini 2.5 Pro, GPT-5, Qwen 3 VL, and o3.}
  \label{fig:example_pred}
\end{figure*}

We first analyze performance across embodiment types for both VLMs and human experts (see \cref{fig:bench_res}).
As a naive uninformed reference, we include the Straight Forward baseline.
Human experts clearly outperform all VLMs, highlighting the gap between model capabilities and task difficulty.
Among the models, Gemini 2.5 Pro ranks best, followed by GPT-5, Qwen 3 VL, and o3 with the Straight Forward baseline ranking unexpectedly close behind o3.
Example predictions of the top four models are shown in \cref{fig:example_pred}.
Generally, we do not observe significant differences between embodiment types for all the models.

Turning to task categories in \cref{fig:bench_res} on the right, we again observe only minor variation.
This uniformity should not be mistaken for balanced competence.
Rather, the overall weakness of the models masks whether category and embodiment-specific differences exist.
The competitiveness of the naive Straight Forward baseline highlights this deficit.

Our experiments demonstrate that Gemini 2.5 Pro achieves the best overall performance in general navigation capabilities.
To gain a clearer understanding of the challenges involved in navigation, we decompose the task into goal-point prediction and path-shape prediction.
Therefore, we compare Gemini 2.5 Pro with baseline models that have access to privileged information as well as with human experts (see \cref{tab:base_res}).
Using Gemini 2.5 Pro to only predict the goal point and then connecting it with a straight line yields only slightly worse results than having Gemini 2.5 Pro predict the full trace.
While baselines with explicit access to the goal point perform significantly better, suggesting that locating the goal area is already a major challenge.
In particular, predicting only a path shape with Gemini 2.5 Pro performs better than Oracle-Goal Straight Line, showing that the model possesses a basic understanding of the scenarios.
However, even when providing the goal point, Gemini 2.5 Pro falls short of human expert performance without this advantage.
Overall, Gemini 2.5 Pro struggles especially in recognizing goal areas but also underperforms in shaping meaningful paths, highlighting the dual difficulty of the task.

\begin{table}[tbp]
\centering
\caption{Comparison between informed baselines, a human and Gemini 2.5 Pro. Note that a higher score is better.} \label{tab:base_res}
\adjustbox{max width=\columnwidth}{
\begin{tabular}{l m{8em}}
\toprule
Model & Score [$\uparrow$] \\
\midrule
Only goal point with Gemini 2.5 Pro & 29.65 \\
Gemini 2.5 Pro & 34.38 \\
Oracle-Goal Straight Line & 51.89 \\
Only path with Gemini 2.5 Pro & 56.55 \\
Human Expert & \bf{75.40} \\
\bottomrule
\end{tabular}
}
\end{table}

\begin{figure}[htbp]
  \centering
  \includegraphics[width=1.0\columnwidth]{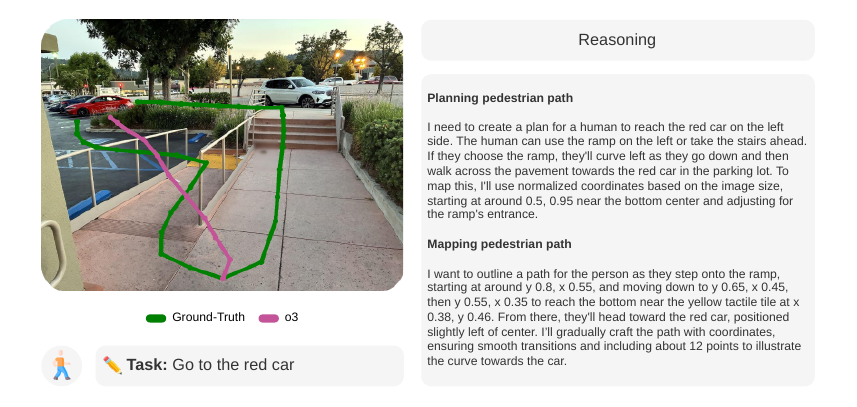}
  \caption{Example of o3's reasoning with the prediction in pink on the left and the steps on the right. The model reasons correctly but is unable to predict a corresponding trace.}
  \label{fig:o3_reasoning}
\end{figure}

Finally, we provide qualitative insights into the reasoning process of models such as Gemini 2.5 Pro and o3.
\cref{fig:o3_reasoning} contains an example reasoning output of o3 where the task is to "go to the red car".
While the model’s textual reasoning correctly distinguishes between the available path options and identifies the correct solution, its predicted trace fails to align with this reasoning.
This is a common pattern we observe when qualitatively analyzing o3's reasoning and suggests a gap between linguistic reasoning and spatial grounding, particularly in localizing traversable structures within the image.

\subsection{Summary of Key Findings}

Our evaluation reveals four critical insights about current VLM navigation capabilities and areas of future work:

\textbf{(1) Large human performance gap.}
Across all four embodiments and task categories, VLM scores are substantially worse than both human and oracle-like baselines, highlighting significant room for improvement (see \cref{fig:bench_res} and \cref{tab:base_res}).
\textbf{(2) Goal localization is the dominant failure mode.}
When models predict only the goal location and we connect it with a straight line, scores are similar to full-trace predictions.
Yet even with the correct goal, path shaping lags behind human performance (see \cref{tab:base_res}).
\textbf{(3) Embodiment robustness.}
Aggregate performance differences across Human, Legged Robot, Wheeled Robot, and Bicycle embodiments are small, suggesting general limitations in spatial grounding rather than embodiment-specific blind spots (see \cref{fig:bench_res}).
\textbf{(4) Score function alignment with human preference.}
Our semantic-aware trace score, that builds on the DTW distance~\cite{dtw} with endpoint error and embodiment-conditioned penalties using automated semantics~\cite{mask2former,mapillary}, correlates more strongly with human preference than DTW alone. 
Using manual segmentation yields an additional but modest gain.

\section{CONCLUSION}\label{sec:conc}

We presented \method, a novel benchmark for evaluating VLM navigation capabilities across different embodiments, along with a novel semantic-aware scoring function for fair evaluation of 2D navigation traces. 
\method provides the first systematic evaluation framework for embodied navigation in real-world scenarios, featuring $1,000$ diverse images from urban and rural environments and four embodiment types.
Our benchmark extends pointing tasks to sequential navigation prediction, creating a natural bridge between high-level VLM reasoning and low-level robotic control. 

To encourage future progress, we will make \method publicly available with test tasks, a leaderboard, and validation split for potential fine-tuning applications.
\method establishes an essential testbed for developing and evaluating navigation-capable VLMs, enabling advances in embodied AI towards truly capable robotic navigation systems.

\section{LIMITATIONS}\label{sec:limit}

\method has several key limitations.
The dataset is geographically concentrated in Switzerland, which may limit generalizability to other regions with different infrastructure and navigation norms.
The benchmark is restricted to single-image scenarios, preventing evaluation of temporal reasoning and multi-step planning required in dynamic environments.
The current embodiment selection is limited to ground vehicles and excludes aerial drones.
Additionally, our semantic scoring function relies on automated segmentation models that may introduce systematic evaluation biases.

While annotating traces has proven to be easy and efficient, the proposed scoring function---although shown to align effectively with human preferences and sufficient for evaluating VLM navigation capabilities---may still fail to capture more nuanced aspects of human preferences.
For instance, while multiple ground-truth traces can capture ambiguities, they constrain the score function to recognize only specific points as goals rather than broader targets such as an entire doorway.
Furthermore, it is not possible to take into account whether a trace works for precisely defined robot dimensions.

\section*{ACKNOWLEDGMENT}
This work was supported by the German Research Foundation (DFG) – 448648559, Luxembourg National Research Fund (Ref. 18990533), and the Swiss National Science Foundation (SNSF) as part of the projects No.200021E\_229503 and No.227617. We thank Kaiqi Qu, Omkar Jarande, and Qicai Tan for joining us in the labeling effort.
We also thank all the people helping us collect images and participating in the evaluation of human performance.

\bibliographystyle{icra/IEEEtran}
\bibliography{bibliography}

\appendix

\subsection{Prompts}

We tested several prompt versions during the development of \method, and the final configuration employs a detailed task description.
In this setup, the system prompt explicitly defines the task, output format, and embodiment behavior, whereas the user prompt specifies the particular navigation scenario that the model must solve.

All navigation trace coordinates are normalized to the interval $[0,1]$ to ensure consistency across images of varying resolutions.
Although models should ideally be robust to the type of normalization applied during training, improvements in performance may occur when using model-specific normalization schemes.

Because the task differs slightly for baselines that predict only a goal point or path to generating a complete trace (see \Cref{sec:baselines}), three distinct prompts are used:
\begin{itemize}
    \item \textbf{Model inference prompt:} \Cref{lst:normal_sys_prompt} (System), \Cref{lst:normal_usr_prompt} (User)
    \item \textbf{Only predict goal point prompt:} \Cref{lst:goal_sys_prompt} (System), \Cref{lst:goal_usr_prompt} (User)
    \item \textbf{Only predict path prompt:} \Cref{lst:path_sys_prompt} (System), \Cref{lst:path_usr_prompt} (User)
\end{itemize}

\begin{lstlisting}[language=prompt, caption={Model inference system prompt}, label={lst:normal_sys_prompt}]
You are a navigation expert for various embodiments including robots and humans. Given an image of the current scenario, a specified embodiment (e.g., legged robot, wheeled robot, human, or bike), and a navigation task (e.g., "Go down the road"), you will predict a feasible future trajectory as a sequence of 2D points in normalized image coordinates (ranging from 0 to 1, where [0,0] is the top-left and [1,1] is the bottom-right).

- The image shows a first-person view of the navigation scenario
- Start your trajectory near the bottom center of the image, which corresponds approximately to normalized coordinate [0.5, 0.95] (representing the current position of the embodiment)
- The trajectory should be adapted to the embodiment's abilities and limitations
- Plan the path forward from this starting position based on what the embodiment can see and navigate
- The trajectory should extend all the way to the goal if the path is visible. If the path is occluded, the trajectory should end where the path becomes fully obscured, unless the path can be reasonably inferred from the visible context.
- If a red traffic light is visible and affects the planned path, or if crossing traffic or moving vehicles are present that make it unsafe to proceed, stop at an appropriate waiting position (e.g., just before the intersection or curb) and end the trajectory there.
- All tasks that you are given have a solution
- Output **only** the list of 2D points in normalized image coordinates (values between 0 and 1) in the following format: `[[x1, y1], [x2, y2], ..., [xn, yn]]`
- Do not include any explanation or additional output

### Embodiment Movement Characteristics

- **Human**: A standard pedestrian. Can navigate stairs and ramps but cannot climb tall obstacles.
- **Legged Robot**: A quadruped like ANYmal. Behaves similarly to a human, but it is shorter. It can handle stairs and escalators.
- **Wheeled Robot**: A wheeled delivery robot. Behaves like a wheelchair, preferring smooth surfaces such as walkways and ramps. It cannot use stairs or escalators.
- **Bicycle**: A standard cyclist. Follows traffic regulations and prefers bike lanes or streets. Cannot navigate stairs.
\end{lstlisting}

\begin{lstlisting}[language=prompt, caption={Model inference user prompt}, label={lst:normal_usr_prompt}]
**Embodiment**: {embodiment}
**Task**: {task}

The image shows a first-person view from the embodiment's current position. Begin your trajectory near the bottom center of the image (around normalized coordinate [0.5, 0.95]) and predict the path forward as a list of 2D points in normalized coordinates (values from 0 to 1) according to the embodiment and the scenario shown in the image.
\end{lstlisting}

\begin{lstlisting}[language=prompt, caption={Only predict goal point system prompt}, label={lst:goal_sys_prompt}]
You are a navigation expert for various embodiments including robots and humans. Given an image of the current scenario, a specified embodiment (e.g., legged robot, wheeled robot, human, or bike), and a navigation task (e.g., "Go down the road"), you will predict a feasible goal point as a 2D point in normalized image coordinates (ranging from 0 to 1, where [0,0] is the top-left and [1,1] is the bottom-right).

- The image shows a first-person view of the navigation scenario
- The predicted goal point should be adapted to the embodiment's abilities and limitations 
- Plan the goal point forward from the embodiment's current position (near the bottom center of the image, which is approximately normalized coordinate [0.5, 0.95]) based on what the embodiment can see and navigate
- If the intended path is occluded, the goal point should be where the path becomes fully obscured, unless the path can be reasonably inferred from the visible context.
- If a red traffic light is visible and affects the intended path, or if crossing traffic or moving vehicles are present that make it unsafe to proceed, place the goal point at an appropriate waiting position (e.g., just before the intersection or curb).
- All tasks that you are given have a solution
- Output **only** the 2D point in normalized image coordinates (values between 0 and 1) in the following format: `[x, y]`
- Do not include any explanation or additional output 

### Embodiment Movement Characteristics

- **Human**: A standard pedestrian. Can navigate stairs and ramps but cannot climb tall obstacles.
- **Legged Robot**: A quadruped like ANYmal. Behaves similarly to a human, but it is shorter. It can handle stairs and escalators.
- **Wheeled Robot**: A wheeled delivery robot. Behaves like a wheelchair, preferring smooth surfaces such as walkways and ramps. It cannot use stairs or escalators.
- **Bicycle**: A standard cyclist. Follows traffic regulations and prefers bike lanes or streets. Cannot navigate stairs.
\end{lstlisting}

\begin{lstlisting}[language=prompt, caption={Only predict goal point user prompt}, label={lst:goal_usr_prompt}]
**Embodiment**: {embodiment}
**Task**: {task}

The image shows a first-person view from the embodiment's current position. Predict a feasible goal point in normalized image coordinates (values from 0 to 1) according to the embodiment and the scenario shown in the image.
\end{lstlisting}

\begin{lstlisting}[language=prompt, caption={Only predict path system prompt}, label={lst:path_sys_prompt}]
You are a navigation expert for various embodiments including robots and humans. Given an image of the current scenario, a specified embodiment (e.g., legged robot, wheeled robot, human, or bike), a navigation task (e.g., "Go down the road"), a starting point, and a goal point, you will predict a feasible future trajectory as a sequence of 2D points in normalized image coordinates (ranging from 0 to 1, where [0,0] is the top-left and [1,1] is the bottom-right).

- The image shows a first-person view of the navigation scenario.
- The trajectory should start at the provided '{start_point}' and aim to reach '{goal_point}'.
- The trajectory should be adapted to the embodiment's abilities and limitations, and consider the overall navigation task.
- Plan the path forward based on what the embodiment can see and navigate.
- All tasks that you are given have a solution
- Output **only** the list of 2D points in normalized image coordinates (values between 0 and 1) in the following format: `[[x1, y1], [x2, y2], ..., [xn, yn]]`
- Do not include any explanation or additional output.

### Embodiment Movement Characteristics

- **Human**: A standard pedestrian. Can navigate stairs and ramps but cannot climb tall obstacles.
- **Legged Robot**: A quadruped like ANYmal. Behaves similarly to a human, but it is shorter. It can handle stairs and escalators.
- **Wheeled Robot**: A wheeled delivery robot. Behaves like a wheelchair, preferring smooth surfaces such as walkways and ramps. It cannot use stairs or escalators.
- **Bicycle**: A standard cyclist. Follows traffic regulations and prefers bike lanes or streets. Cannot navigate stairs.
\end{lstlisting}

\begin{lstlisting}[language=prompt, caption={Only predict path user prompt}, label={lst:path_usr_prompt}]
**Embodiment**: {embodiment}
**Task**: {task}
**Start Point**: {start_point}
**Goal Point**: {goal_point}

The image shows a first-person view. Predict the path from the **Start Point** to the **Goal Point** as a list of 2D points in normalized coordinates (values from 0 to 1) according to the embodiment, the specified task, and the scenario shown in the image.
\end{lstlisting}

\subsection{Semantic Penalty}

In the score function (see \Cref{sec:score}), semantic penalties are assigned to image regions based on their predicted class labels obtained from Mask2Former. 
For each embodiment $e$, we define a manually specified mapping from semantic class $S_i$ to a penalty value $\textrm{m}_{e}(S_i)$. 
These embodiment-dependent penalty values are listed in \Cref{app:semantic_map}.

\begin{table}[htbp]
\centering
\caption{Manually calibrated semantic penalties $\textrm{m}_{e}(S_i)$ for each class $S_i$ and embodiment $e$.} \label{app:semantic_map}
\adjustbox{max width=0.81\columnwidth}{
\begin{tabular}{p{11em} r r r r}
\toprule
Class & Human & Legged Robot & Wheeled Robot & Bicycle \\
\midrule
Bird & 0 & 0 & 0 & 0 \\
Ground Animal & 200 & 200 & 300 & 600 \\
Curb & 200 & 200 & 300 & 300 \\
Fence & 600 & 600 & 600 & 600 \\
Guard Rail & 600 & 600 & 600 & 600 \\
Barrier & 800 & 800 & 800 & 800 \\
Wall & 1000 & 1000 & 1000 & 1000 \\
Bike Lane & 500 & 500 & 500 & 0 \\
Crosswalk - Plain & 0 & 0 & 0 & 0 \\
Curb Cut & 0 & 0 & 0 & 0 \\
Parking & 400 & 400 & 400 & 400 \\
Pedestrian Area & 0 & 0 & 0 & 0 \\
Rail Track & 400 & 400 & 600 & 600 \\
Road & 400 & 400 & 400 & 0 \\
Service Lane & 400 & 400 & 400 & 0 \\
Sidewalk & 0 & 0 & 0 & 400 \\
Bridge & 0 & 0 & 0 & 0 \\
Building & 1000 & 1000 & 1000 & 1000 \\
Tunnel & 0 & 0 & 0 & 0 \\
Person & 200 & 200 & 200 & 400 \\
Bicyclist & 200 & 200 & 200 & 200 \\
Motorcyclist & 500 & 500 & 500 & 500 \\
Other Rider & 500 & 500 & 500 & 500 \\
Lane Marking - Crosswalk & 0 & 0 & 0 & 0 \\
Lane Marking - General & 0 & 0 & 0 & 0 \\
Mountain & 1000 & 1000 & 1000 & 1000 \\
Sand & 100 & 100 & 500 & 500 \\
Sky & 1000 & 1000 & 1000 & 1000 \\
Snow & 150 & 150 & 800 & 800 \\
Terrain & 800 & 800 & 800 & 800 \\
Vegetation & 800 & 800 & 800 & 800 \\
Water & 1000 & 1000 & 1000 & 1000 \\
Banner & 800 & 800 & 800 & 800 \\
Bench & 300 & 300 & 300 & 300 \\
Bike Rack & 500 & 500 & 500 & 500 \\
Billboard & 800 & 800 & 800 & 800 \\
Catch Basin & 100 & 100 & 200 & 200 \\
CCTV Camera & 0 & 0 & 0 & 0 \\
Fire Hydrant & 500 & 500 & 500 & 500 \\
Junction Box & 500 & 500 & 500 & 500 \\
Mailbox & 300 & 300 & 300 & 300 \\
Manhole & 0 & 0 & 0 & 0 \\
Phone Booth & 400 & 400 & 400 & 400 \\
Pothole & 300 & 300 & 300 & 600 \\
Street Light & 300 & 300 & 300 & 300 \\
Pole & 300 & 300 & 300 & 300 \\
Traffic Sign Frame & 300 & 300 & 300 & 300 \\
Utility Pole & 300 & 300 & 300 & 300 \\
Traffic Light & 300 & 300 & 300 & 300 \\
Traffic Sign (Back) & 500 & 500 & 500 & 500 \\
Traffic Sign (Front) & 500 & 500 & 500 & 500 \\
Trash Can & 400 & 400 & 400 & 400 \\
Bicycle & 400 & 400 & 400 & 400 \\
Boat & 500 & 500 & 500 & 500 \\
Bus & 800 & 800 & 800 & 800 \\
Car & 800 & 800 & 800 & 800 \\
Caravan & 800 & 800 & 800 & 800 \\
Motorcycle & 800 & 800 & 800 & 800 \\
On Rails & 800 & 800 & 800 & 800 \\
Other Vehicle & 800 & 800 & 800 & 800 \\
Trailer & 800 & 800 & 800 & 800 \\
Truck & 800 & 800 & 800 & 800 \\
Wheeled Slow & 800 & 800 & 800 & 800 \\
Car Mount & 800 & 800 & 800 & 800 \\
Ego Vehicle & 800 & 800 & 800 & 800 \\
\bottomrule
\end{tabular}
}
\end{table}

\subsection{Model Configuration}

Experiments are conducted using API-based access to all models.
Default provider hyperparameters are used, except for a maximum token limit of $5000$ and a temperature of $1.0$, which are consistently applied across models.
We give no explicit instructions to produce intermediate reasoning steps.
Models capable of generating such steps do so according to their default behavior.
\Cref{app:models} summarizes the key characteristics of the considered models.

\begin{table}[htbp]
\centering
\caption{Overview of evaluated models and selected properties.} \label{app:models}
\adjustbox{max width=0.81\columnwidth}{
\begin{tabular}{p{8em} c c c}
\toprule
Model & Number of Parameters & Open-Weight & Reasoning Steps \\
\midrule
Gemini 2.5 Pro & \textit{N/A} & \redcross & \greencheck \\
GPT-5 & \textit{N/A} & \redcross & \greencheck \\
o3 & \textit{N/A} & \redcross & \greencheck \\
Claude Sonnet 4 & \textit{N/A} & \redcross & \redcross \\
Mistral Medium 3.1 & \textit{N/A} & \redcross & \redcross \\
Qwen 3 VL & 235B A22B & \greencheck & \greencheck \\
Qwen 2.5 VL & 72B & \greencheck & \redcross \\
Gemma 3 & 27B & \greencheck & \redcross \\
\bottomrule
\end{tabular}
}
\end{table}

\end{document}